\documentclass[sigplan,,nonacm,screen]{acmart}
\usepackage{multirow}
\usepackage[most]{tcolorbox}
\usepackage{enumitem}
\usepackage{graphicx}
\usepackage{subcaption}
\usepackage{caption}
\usepackage{algorithm}
\usepackage{algpseudocode}
\usepackage{float}

\newtcolorbox{takeaway}[1][]{
  colback=gray!6!white,
  colframe=gray!50!black,
  fonttitle=\bfseries,
  coltitle=black,
  left=3pt,right=3pt,top=2pt,bottom=2pt,
  boxsep=1pt,
  before skip=4pt,
  after skip=4pt,
  enhanced,
  breakable,
  arc=4pt,
  boxrule=0.5pt,
  #1
}

\AtBeginDocument{%
  }

\newcommand*\circled[1]{\tikz[baseline=(char.base)]{
            \node[shape=circle,fill,inner sep=0.2pt] (char) {\textcolor{white}{#1}};}}

\author{Ritik Raj}
\affiliation{%
  \institution{Georgia Institute of Technology}
  \country{USA}
}
\authornote{Corresponding authors: Ritik Raj (ritik.raj@gatech.edu) and Souvik Kundu (souvikk.kundu@intel.com)}

\author{Souvik Kundu}
\affiliation{%
  \institution{Intel}
  \country{USA}
}
\authornotemark[1]

\author{Ishita Vohra}
\affiliation{%
  \institution{Georgia Institute of Technology}
  \country{USA}
}

\author{Hong Wang}
\affiliation{%
  \institution{Intel}
  \country{USA}
}

\author{Tushar Krishna}
\affiliation{%
  \institution{Georgia Institute of Technology}
  \country{USA}
}

\begin{document}

\title[]{Towards Understanding, Analyzing, and Optimizing Agentic AI Execution: A CPU-Centric Perspective}

\begin{abstract}
Agentic AI serving converts monolithic LLM-based inference to autonomous problem-solvers that can plan, call tools, perform reasoning, and adapt on the fly. Due to diverse task execution need, such serving heavily rely on heterogeneous CPU–GPU systems with majority of the external tools responsible for agentic capability, either run on or are orchestrated by the CPU.
 Towards having a deeper understanding of its role, this paper aims to characterize and analyze the system bottlenecks introduced by agentic AI workloads from a largely overlooked CPU-centric perspective. We first present a compile-time characterization of agentic AI execution and choose representative workloads to capture the algorithmic diversity. We then perform runtime characterization of the representative workloads analyzing the end-to-end latency and throughput on two different hardware systems to isolate respective architectural bottlenecks. Based on the insights on the bottlenecks, we finally present two scheduling optimizations, namely,  \circled{1} \textbf{C}PU-Aware \textbf{O}verlapped \textbf{M}icro-\textbf{B}atching (\textbf{COMB}) and \circled{2} \textbf{M}ixed \textbf{A}gentic \textbf{S}cheduling (\textbf{MAS}) on homogeneous and heterogeneous agentic workloads\footnote{In this work, we refer to homogeneous agentic workload as single agentic workload type (e.g., CPU-heavy)  while heterogeneous workload refers to a mix of two agentic workload types (CPU-heavy and GPU-heavy).}, respectively. In specific, these methods optimize for improved CPU-GPU concurrent utilization while reducing skewed resource allocation for heterogeneous execution. Experimental evaluations on the two hardware systems demonstrate the efficacy of COMB in yielding up to \textbf{1.7$\times$} lower P50 latency in standalone homogeneous workload execution and up to \textbf{3.9$\times$/1.8$\times$} lower service/total latency under homogeneous open-loop load. Additionally, for heterogeneous open-loop load, MAS can reduce the total latency for minority request-type by up to \textbf{2.37$\times$/2.49$\times$} at P50/P90 percentile.

\end{abstract}

\maketitle

\section{Introduction}

Large Language Models (LLMs) have spearheaded the advancements in Artificial Intelligence (AI) for a plethora of applications, including vision \cite{wang2023visionllm, zhou2024navgpt}, healthcare \cite{thirunavukarasu2023large, bedi2025testing}, science \cite{telenti2024large, jablonka2024leveraging}, and education \cite{gan2023large, wang2024large}. However, they face challenges including context-agnosticism \cite{berglund2023taken}, hallucinations \cite{maynez2020faithfulness} and the lack of real-time information \cite{komeili2021internet, ouyang2023enhancing}. These challenges have fueled the emergence of \textit{agentic AI systems}, where LLMs interact with external tools to gain agency beyond the standalone intelligence of monolithic LLMs.

\begin{figure}[!t]
    \includegraphics[width=\columnwidth]
    {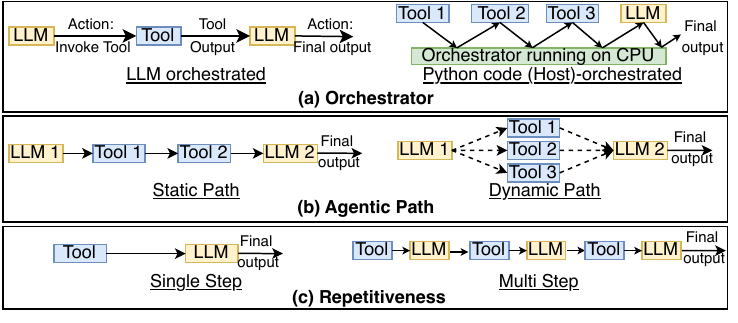}
    \vspace{-6mm}
    \caption{Compile-time Characterization of agentic AI on the basis of (a) Orchestrator (LLM/Host) (b) Agentic Path (Static/Dynamic) and (c) Repetitiveness (Single/Multi-step).}
    \vspace{-3mm}
    \label{fig:characterization}
\end{figure}

\begin{table}[!t]
\centering
\footnotesize
\caption{CPU, GPU and memory specifications of two different systems used for characterization and evaluation.}
\label{tab:system_types}
\begin{tabular}{l|l|l}
\hline
Component           & \texttt{Sys 1}: HP CPU, LP GPU                         & \texttt{Sys 2}: HP CPU, HP GPU   \\ \hline
CPU        & \begin{tabular}[c]{@{}c@{}} 64-core Intel GNR \end{tabular} & 72-core Nvidia Grace \\ \hline
CPU Memory & DDR5 512 GB                       & LPDDR5 480 GB \\ \hline
GPU        & Nvidia-RTX-Pro 6000                           & Nvidia H200  \\ \hline
GPU Memory & GDDR7 96 GB                  & HBM3e 96 GB  \\ \hline

\end{tabular}
\end{table}

\begin{table*}[t]
\small
    \centering
    \caption{Compile-time characterization of representative workloads. Tools/Application considered for profiling are underlined.}
    \begin{tabular}{c|c|c|c|c|c}
    \hline
       \multirow{2}{*}{\textbf{Agentic Workload}} & \multicolumn{3}{|c|}{\textbf{Compile-time Characterization}} & \multirow{2}{*}{\textbf{Tools}} & \multirow{2}{*}{\textbf{Application}} \\
       \cline{2-4}
       & \textbf{Orchestrator} & \textbf{Path} & \textbf{Flow}  & & \\
       \hline
       \textbf{Toolformer \cite{schick2023toolformer}} & LLM & Dynamic & Single-step & \underline{Calculator API},  Calendar & MLQA, \underline{Math}  \\
            \hline

    \textbf{SWE-Agent \cite{yang2024swe}} & LLM & Static & Multi-step & \underline{Bash (File I/O and Python Execution)} & \underline{SDE},  Data analysis \\

    \hline

    \textbf{RAG (Haystack \cite{deepset-haystack})} & Python code (Host)  & Static & Single-step & Web search, \underline{Document Retrieval} & \underline{RAG QA} \\
            \hline

        \textbf{ChemCrow \cite{bran2023chemcrow}} & LLM & Dynamic & Multi-step & \underline{Conformer Gen tool}, Reaction Tools &  \underline{Chemistry Research} \\
\hline
        
   \begin{tabular}[c]{@{}c@{}} \textbf{Web-Augmented} \\ \textbf{Agent (LangChain\cite{mavroudis2024langchain})} \\ \end{tabular}  & Python code (Host) & Static &  Single-step &  \underline{Web search}, \underline{summarizer} & \begin{tabular}[c]{@{}c@{}} \underline{Web-based QA}, \\ DevOps\end{tabular} \\
            \hline

    \end{tabular}
    \vspace{-1em}
    
    \label{tab:characterization}
\end{table*}

Agentic AI frameworks \cite{schick2023toolformer, singh2025agentic} orchestrate multiple components including tool use, memory modules, and iterative reasoning loops to achieve superior performance compared to monolithic LLMs. Recent benchmarks reveal that agentic frameworks such as ReAct \cite{yao2023react} achieve 34\% higher success rates on ALFWorld \cite{shridhar2020alfworld} tasks and 10\% improvement on WebShop \cite{yao2022webshop} compared to equivalent-sized monolithic models, while AutoGPT \cite{yang2023auto} and BabyAGI \cite{nakajimababyagi} demonstrate up to 3$\times$ better performance on long-horizon planning tasks despite using smaller base models. The performance advantages are particularly pronounced for domains requiring external knowledge integration and iterative refinement. For example, WebGPT \cite{nakano2021webgpt} shows that 7B parameter models can match or outperform 70B monolithic models on knowledge-intensive tasks, while  achieving 64.1\% accuracy on TruthfulQA \cite{lin2021truthfulqa} compared to 59.3\% for GPT-3 \cite{brown2020language} despite being 25$\times$ smaller.

Although  AI models run mostly on GPUs, CPUs are used in tool processing including Bash execution, web search, lexical summarization \cite{erkan2004lexrank}, and Exact Nearest Neighbor Search (ENNS) on large databases. While prior approaches on AI efficiency aggressively focused on GPU kernels and KV-cache management \cite{kwon2023efficient}, they become ineffective for the CPU-centric tool execution of the agentic AI workloads. A recent work \cite{quinn2025accelerating} shows that ENNS accounts for more than 75\% of the end-to-end (E2E) latency on a 200 GB document corpus for a Retrieval Augmented Generation (RAG) workload with a Llama-3-70B \cite{dubey2024llama}. %
Furthermore, \cite{patel2024large} argued that web agent benchmarks like WebArena \cite{zhou2023webarena} are computationally intensive due to latency from real-time web interactions, where LLM actions can't be batched. \cite{xu2024conveyor} shows that partial tool execution can cut request latency by up to 38.8\%, highlighting tool execution as a major source of E2E latency.

\noindent
\textbf{Our Contributions.} To address this emergent CPU bottleneck, this work presents a two-fold contribution. \underline{Firstly, } we present a \textit{compile-time and runtime characterization} to understand the system implications of CPU-centric (or tool-centric) agentic workloads. \underline{Secondly,} we present \textit{scheduling optimization solutions} to essentially improve concurrent CPU-GPU utilization in agentic AI serving systems.

\textit{Compile-time and Runtime Characterization:} We first introduce a compile-time characterization (\autoref{section:characterization})  by selecting representative workloads to comprehensively capture the algorithmic and computational diversity of agentic AI. To benchmark, we categorize based on three metrics, namely, \textit{orchestration}, \textit{agentic path type}, and \textit{task repetitiveness} as shown in \autoref{fig:characterization}. We then conduct an in-depth runtime characterization on different hardware systems through end-to-end latency (\autoref{subsection: latency profiling}), batch throughput (\autoref{subsection: throughput}) and energy profiling (\autoref{ablation:energy_profiling}) to isolate the major hardware bottlenecks (CPU, GPU or I/O) specific to the system. In specific, we perform the experiments on two different CPU-GPU settings as shown in \autoref{tab:system_types} with relative high-performance (HP) and low-performance (LP) GPU counterparts. Interestingly, we find that tool dominated agentic AI workloads are significantly bottle-necked by tool processing on the CPU consuming up to 88\% of the end-to-end latency. With better quality of GPUs, the bottleneck can swiftly shift more towards CPUs. More importantly, CPU-parallelization strategies often exhibit lower efficiency than their GPU counterparts, prematurely saturating the throughput that can reduce the GPU utilization. \textbf{This necessitates the CPU execution to be carefully optimized to improve the execution latency for agentic workloads}.

\textit{Scheduling Optimizations:} Based on the throughput saturation insights, we present two scheduling optimizations for agentic workloads. In particular, for homogeneous workloads, to avoid premature saturation of throughput, we present CPU-Aware Overlapped Micro-Batching (COMB - \autoref{subsection:comb}). On the other hand, for heterogeneous workloads, we propose a novel scheduling policy dubbed as Mixed Agentic Scheduling (MAS - \autoref{subsection:mas}) to maintain fair utilization of both CPU-GPU resources and improve performance during real-server-like bursty arrival patterns. In specific, these methods optimize for improved CPU-GPU concurrent utilization while reducing skewed resource allocation for
heterogeneous execution. We showcase the generalization of the proposed optimizations on two different hardware platforms. COMB shapes homogeneous request-type concurrency and improves CPU-GPU utilization, yielding up to $3.9\times$ lower service latency, $1.8\times$ lower total latency, and $1.7\times$ higher throughput under open-loop load. MAS, on the other hand, protects the minority request-type under mixed CPU/GPU workloads, improving $P50/P90$ latency by up to $2.37\times/2.49\times$.
To the best of our knowledge, this is the first work to quantify and analyze end-to-end latency, throughput, and energy bottlenecks in agentic AI execution for heterogeneous CPU-GPU systems. We believe this work will inspire the next-frontier of agentic AI serving systems to have the optimal concurrent CPU-GPU utilization as a key design principle.

\section{Compile-time Characterization}
\label{section:characterization}

Prior work has largely categorized agentic AI through the lens of agent capabilities. For instance, a recent study \cite{sapkota2025ai} contrasts agentic AI systems, characterized by distributed cognition, persistent memory, and coordinated planning, with traditional single-agent systems oriented toward task-specific automation. On the contrary, we introduce three orthogonal bases as shown in \autoref{fig:characterization} for classifying agentic AI that directly influence algorithmic and system-level metrics. This taxonomy is intended to serve as a priori, compile-time platform-agnostic characterization.

\subsection{Three Orthogonal Classification of Agentic AI}
\underline{First}, on the basis of the \textit{orchestrator}, we divide agentic AI systems into LLM-orchestrated and host-orchestrated (through Python code). In the LLM-orchestrated agentic AI workloads, the LLM controls the end-to-end execution flow. In the pipeline, the LLM, working as an orchestrator, decides whether to invoke the tool or emit final output. On the other hand, host orchestrated workloads call host/python code to determine the next agent (tool/LLM) in the pipeline. \underline{Second}, on the basis of the \textit{agentic path}, we divide agentic AI systems as static-path and dynamic-path systems. Static-path agentic systems follow a predetermined path while dynamic-path systems determine the path during runtime based on the orchestrator. In other words, the orchestrator has path decision making capability for dynamic-path agentic systems. For static-path systems, the orchestrator is only responsible for communication between different agents in the pipeline. \underline{Third}, on the basis of the \textit{repetitiveness}, we divide agentic AI into single-step and multi-step systems. Single-step agentic systems are more prevalent in standalone web-based or RAG-based retrieval execution where single call to these tools is sufficient to complete the task. while multi-step systems are more prevalent in gaming, robotics or similar applications that require multiple interactions to execute the task.

\subsubsection{Orchestrator-based Classification} This dimension characterizes systems based on where the primary orchestration logic resides. LLM-orchestrated systems delegate control flow decisions to the language model itself, leveraging its reasoning capabilities for task decomposition and execution planning. In contrast, Python code (host)-orchestrated systems employ traditional programmatic control structures, with the CPU managing task scheduling, tool invocation, and result aggregation while treating the LLM as a stateless inference engine. Examples are as follows:

\textbf{LLM-orchestrated}: ReAct \cite{yao2023react}, AutoGPT \cite{yang2023auto}, BabyAGI \cite{nakajimababyagi}, AgentGPT \cite{agentgpt}, CAMEL \cite{li2023camel}, MetaGPT \cite{hong2024metagpt}

\textbf{Python code (Host)-orchestrated}: LangChain \cite{mavroudis2024langchain}, Semantic Kernel \cite{microsoft-no-date}, Haystack \cite{deepset-haystack}, LlamaIndex \cite{llamaindex}

\subsubsection{Path-based Classification}
This dimension distinguishes between predetermined and adaptive execution strategies. Static-path agents follow predefined workflows with deterministic tool invocations. Dynamic-path agents adaptively construct execution graphs based on intermediate results, environmental feedback, and emergent task requirements. 

\textbf{Static-Path:} Haystack \cite{deepset-haystack}, LlamaIndex \cite{llamaindex}

\textbf{Dynamic-Path:}  Reflexion \cite{shinn2023reflexion}, LATS \cite{zhou2023language}

\subsubsection{Flow/Repetitiveness-based Classification}

This taxonomy captures the iterative nature of agent-environment interactions. Single-step agents complete tasks in a single inference pass without environmental feedback. Multi-step repetitive agents engage in iterative refinement cycles for complex tasks requiring extensive exploration.

\textbf{Single-step:} CoT prompting systems, Zero-shot tool use, Single-turn QA agents, RAG \cite{lewis2020retrieval}

\textbf{Multi-step:} WebArena \cite{zhou2023webarena}, Balrog \cite{paglieri2024balrog}, AgentBench \cite{liu2023agentbench}

\subsection{Representative Workloads}
\label{sec:selected_workloads}
\subsubsection{Workload Overview}
\label{subsec:model_overview}
We select five agentic AI workloads for profiling analysis as detailed in \autoref{tab:characterization}. We evaluate \textit{Toolformer} \cite{schick2023toolformer} on math benchmarks using WolframAlpha API \cite{wolframalpha-instant-calculators-api}, \textit{SWE-Agent} \cite{yang2024swe} on coding benchmarks using Bash execution tool, \textit{ChemCrow} \cite{bran2023chemcrow} on molecular benchmarks using RDKit conformer generation tool \cite{greg_landrum_2026_19250388}, \textit{RAG} implemented via Haystack \cite{deepset-haystack} on Question Answering (QA) benchmarks using ENNS retrieval tool on 115 GB C4 document corpus \cite{dodge2021documenting}, and \textit{Web-Augmented Agent} implemented via LangChain \cite{mavroudis2024langchain} on QA benchmarks leveraging web search and lexical summarization tools. Notably, the Web-Augmented Agent task (web search $\to$ summarization $\to$ LLM inference) is formulated inspired by the web search feature of popular chatbots \cite{gemini2026, chatgpt2026}. In our experiments, we chose a CPU-based lexical summarizer (LexRank \cite{erkan2004lexrank}) instead of an LLM-based summarizer. The lexical summarizer helps reduce hallucination \cite{maynez2020faithfulness} while improving the domain accuracy \cite{giarelis2023abstractive}. Refer to \autoref{appendix: representative workloads} for more workload details.

We select these agentic AI workloads because they are representative of different categories of compile-time characterization as well as applications and tools.  \textit{\underline{Firstly}, challenging applications}: they target factual, coding, and scientific tasks as well as live-data queries where standard LLMs under-perform. \textit{\underline{Secondly}, diverse computational patterns}: these models span a wide range of model sizes, orchestration patterns and tool integration strategies that are representative of broader agentic AI systems. \textit{\underline{Finally}}, these tools are representative of general processing aspects of the CPU. For example, Python execution pipeline of compute-intensive benchmarks in the SWE-Agent tests out the execution units of the CPU.

\begin{figure*}[!t]
    \centering
    \includegraphics[width=0.9\linewidth]{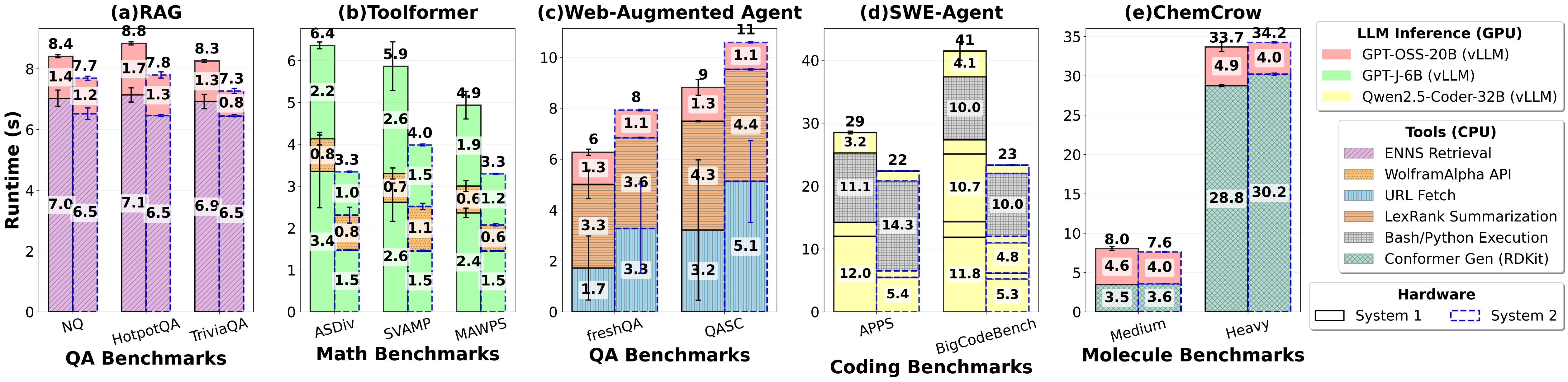}
    \vspace{-1em}
    \caption{(a) End-to-end (E2E) latency for RAG (Haystack); (b) Toolformer; (c) Web-Augmented Agent (LangChain); (d) Mini-SWE-Agent; and (e) ChemCrow on the two different hardware systems (refer to \autoref{tab:system_types}).}
    \vspace{-1em}
    \label{fig:latency}
\end{figure*}

\subsubsection{SLMs for Representative Workloads}
Small Language Models (SLMs) are a good fit for agentic AI \cite{belcak2025smalllanguagemodelsfuture} because agents thrive on fast, iterative perceive-plan-act loops, and privacy-preserving local execution. Many agent competencies are externalized: tool use and retrieval can offload computation and factual recall, reducing reliance on parametric capacity while preserving task performance, a setting in which SLMs including GPT-J 6B \cite{gpt-j} can outperform larger monolithic LLMs including OPT 66B \cite{zhang2022opt} and GPT-3 175B \cite{brown2020language} as shown in \cite{schick2023toolformer}.  Furthermore, recent studies \cite{gunasekar2023textbooks, abdin2024phi} show sub-10B models achieving competitive capability on MMLU \cite{hendrycks2020measuring} and MT-bench \cite{zheng2023judging} benchmarks as compared to GPT-3.5 when trained with high-quality data and efficient architectures. Therefore, in this work, we focus on models having up to 32 B parameters.

\section{Profiling}
\label{section: profiling}

\subsection{System and Software Setup}
\label{subsection: profiling setup}
The experiments are performed on two hardware platforms with asymmetric GPUs to isolate CPU-centric architectural bottlenecks. The first system (\texttt{Sys 1}) consists of Intel 6th generation Xeon Granite Rapids (GNR) CPU (HP) and Nvidia RTX-Pro 6000 Blackwell GPU (LP). On the other hand, the second system (\texttt{Sys 2}) is a GH200 gracehopper system with Nvidia Grace CPU (HP) and H200 GPU (HP). The specifications are summarized in \autoref{tab:system_types}. Our software environment includes PyTorch (version 2.8.0) and a local vLLM server (version 0.14.0) for LLM inference.  We run each workload five times to account for statistical variance.

\subsection{End-to-End (E2E) Latency Analysis}
\label{subsection: latency profiling}
\autoref{fig:latency} profiles E2E runtime latency for the five representative agentic workloads, on the two hardware systems to isolate the architectural bottleneck. %
\subsubsection{Runtime Characterization} 

For QA with RAG \\ (Haystack), ENNS retrieval is the main bottleneck consuming 83\%, 81\% and 82\% of total latency for NQ \cite{kwiatkowski2019natural}, HotpotQA \cite{yang2018hotpotqa}, and TriviaQA \cite{joshi2017triviaqa}, respectively, on \texttt{Sys 1}. On the other hand, ENNS retrieval consumes up to 89\% of total latency on \texttt{Sys 2}. For Toolformer, LLM inference is the main bottleneck consuming $\sim $88\% of total latency for \texttt{Sys 1}. Due to better GPU, LLM inference is much faster on \texttt{Sys 2} reducing the inference delay to 77\% of the total latency. 
Web-Augmented Agent (LangChain) shows huge variation in the URL fetch stage due to the network usage. LexRank summarization tool execution accounts for 55\% and 48\% for freshQA \cite{vu2023freshllms} and QASC \cite{khot2020qasc} benchmarks, respectively, on \texttt{Sys 1}. Similarly, it takes 40-45\% on \texttt{Sys 2} as well. Without web I/O variance during URL fetching, if we just consider the summarization and inference stages, the E2E latency of \texttt{Sys 1} remains similar to that with \texttt{Sys 2}. These results highlight that constraining the number of websites to fetch can yield faster E2E latency as opposed to optimizing the inference model. For ChemCrow workload, we see the conformer generation using RDKit tool dominating E2E latency for heavy molecules (85\% and 88\% on \texttt{Sys 1} and \texttt{Sys 2}, respectively) resulting in similar performance for both the systems. On the other hand, for medium molecules, LLM inference part dominates (58\% and 53\% on \texttt{Sys 1} and \texttt{Sys 2}, respectively).

\begin{takeaway}
    \textbf{Key Takeaway 1: } Tool processing on CPUs can take significant chunk of E2E latency, motivating a CPU-centric optimization strategy. Moreover, a system with HP CPU and LP GPU can match a system with HP GPU in E2E latency on such tool-dominated agentic AI workloads motivating cost-effective agentic AI deployments.
\end{takeaway}

For SWE-Agent, Bash/Python execution accounts for 38\% and 25\% of E2E latency for APPS \cite{hendrycks2021measuring} and BigCodeBench \cite{zhuo2024bigcodebench} benchmarks, respectively, on \texttt{Sys 1}. On the other hand, they account for up to 65\% of the E2E latency on \texttt{Sys 2}. This hints at the highly optimized LLM inference on HP GPU of \texttt{Sys 2} that forces the bottleneck more towards tool execution on the CPU. This is further affirmed by the LLM execution latency reduction from \texttt{Sys 1} to \texttt{Sys 2}. For example, LLM inference bottleneck reduced from 88\% to 77\% in Toolformer workload as we move to \texttt{Sys 2}. %

\begin{takeaway}
    \textbf{Key Takeaway 2: } HP GPU system can shift the bottleneck from GPU to CPU when tool execution latency is comparable to LLM inference latency, making them more CPU-bounded than systems with LP GPU, motivating system-aware optimization strategies. 
\end{takeaway}

\subsection{Throughput Analysis}
\label{subsection: throughput}

\subsubsection{GPU Throughput Analysis.}
\label{subsubsection: throughput_saturation_gpu}
We first assume a hypothetical scenario of GPU-only LLM inference, to disentangle the throughput performance of the GPU. We measure the vLLM GPU throughput as $((BS \times (T_{in}+T_{out}))/{t_{sec}})$, where $BS$ represents the batch-size, with $T_{in}$ and $T_{out}$ representing the total input and output tokens, respectively. ${t_{sec}}$ represents the total time in generating all the tokens across batches. As shown in \hyperref[fig:throughput]{Figure 3a}, the throughput increases steadily with increase in $BS$, confirming that the GPU efficiently exploits the additional parallelism exposed by larger batches. The gains are especially pronounced at moderate batch sizes, where batching improves device utilization and amortizes execution overheads. On the other hand, for large $BS$, the rate of increase of throughput reduces and begins to saturate, particularly, for longer input/output sequences. This trend is consistent with a memory-system bottleneck: as batch size grows, the KV cache footprint scales with the total number of processed tokens, increasing pressure on GPU memory capacity and bandwidth. Notably, although mechanisms such as PagedAttention \cite{kwon2023efficient} reduce memory fragmentation and improve serving efficiency, \textbf{they do not eliminate the underlying capacity and bandwidth limits of GPU memory}.

\subsubsection{Workload Throughput Analysis.}
\label{subsubsection: worloadthroughput}
\hfill\break
\noindent\textbf{CPU Parallelism Choice for Agentic Workloads.}
We analyze the tradeoff between multi-processing (MP) and multi-threading (MT) CPU parallelism strategies. MT has lower memory usage as all the threads share the same memory. On the other hand, MP requires independent memory for each process. Since ENNS retrieval has very high memory usage, we use MT for the RAG (Haystack) workload. MT approach is lightweight and incurs lower creation and switching overhead compared to that with MP. As a result, MT approach works better for I/O workloads. Therefore, we select MT for Toolformer as it contains an I/O tool, i.e. the WolframAlpha API. For CPU-compute intensive tools including LexRank Summarization, Bash/Python execution, and RDKit Conformer generation, MT is ineffective due to Python Global Interpreter Lock (GIL) limitation and could not attain true multi-core performance. Therefore, we choose MP approach for Web-Augmented Agent (LangChain), SWE-Agent, and ChemCrow workloads. We further quantify the GIL bottleneck of MT by comparing it with MP approach for Web-Augmented Agent on \texttt{Sys 2} in \autoref{appendix:section-multiprocessing}. Notably, the CPU throughput on multi-core systems can saturate well before all cores are busy. For instance, a study \cite{balay2019petsc} shows that a dual-socket Haswell node reaches $>$80\% of peak bandwidth on the STREAM benchmark \cite{mccalpin2006stream} with only four processes per socket. If we increase the number of parallel processes beyond the available cores (over-subscription \cite{iancu2010oversubscription}), OS scheduler contention and context switching overheads dominate.

\begin{figure*}
    \centering
    \includegraphics[width=1\linewidth]{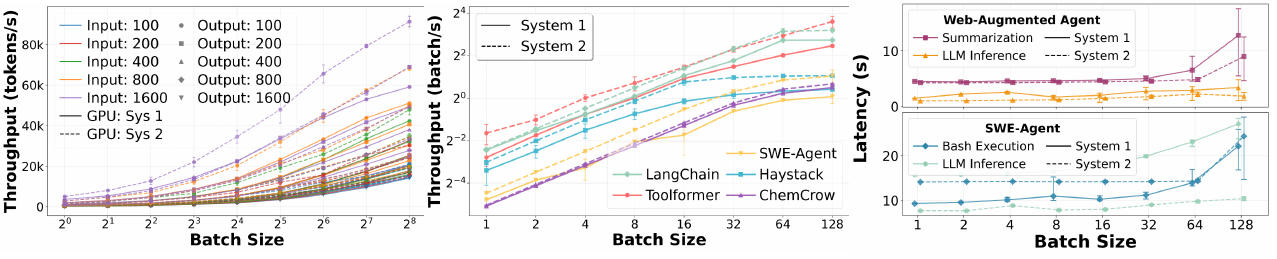}
    \vspace{-3mm}
    \caption{(a) vLLM throughput analysis for GPT-OSS-20B across different batch-sizes with different input-output token lengths; (b) Throughput saturation analysis for various agentic workloads; (c) Average time taken by Web-Augmented Agent and SWE-Agent workload components reveals a critical CPU over-subscription bottleneck at batch size 128 for both the systems.}
    \label{fig:throughput}
    \vspace{-3mm}
\end{figure*}

\noindent
\textbf{Runtime Throughput Analysis.}
We define the throughput of the system on agentic workload as $BS/t_{sec}$.
 \hyperref[fig:throughput]{Figure 3b} demonstrates the throughput variation of representative workloads with batch size ($BS$) scaling. We parallelize each component of the agentic workload including LLM inference on GPU using vLLM and tool processing on CPU using either MP or MT. We showcase different scenarios of throughput boundedness on the five workloads with MAWPS, NQ, QASC, APPS, and large molecule benchmarks, respectively, on the two systems. From this point onward, for the Web-Augmented Agent, we consider only the web independent components (summarization and LLM inference) by substituting URL fetching by on-device cached HTML files.

 For Toolformer, we see the rate of throughput improvement keeps slowing down from 1.9$\times$ to 1.4$\times$ as we move from $BS:$ $1 \to 2$ to $BS:$ $64 \to 128$, for \texttt{Sys 2}. The WolframAlpha API calls are parallelized with nearly zero latency overheads. However, the increased KV cache adds to the throughput saturation of the GPU. For Haystack RAG workload, retrieval is bottle-necked beyond $BS$ = 16/32, for both the systems due to LLC pressure and disk I/O contention arising out of the huge size of the C4 documents. For Web-Augmented Agent (LangChain), SWE-Agent and ChemCrow workloads, the throughput saturates at $BS$ = 128, due to core over-subscription for CPU-heavy tools. \hyperref[fig:throughput]{Figure 3c} further shows that the impact of over-subscription in Web-Augmented Agent (LangChain) and SWE-Agent workloads using average, minimum and maximum time per tool call and LLM inference. The H200 GPU outperforms RTX-6000 Pro Blackwell GPU by 1.9$\times$ and 2.8$\times$, respectively for Web-Augmented Agent and SWE-Agent workloads at $BS=$128. Moreover, the average latency of summarization stage increases by 2.0$\times$ and 1.9$\times$ respectively for \texttt{Sys 1} and \texttt{Sys 2} from $BS=$ 64-128. On the other hand, the average LLM inference latency remained relatively similar for both the hardware platforms from $BS=$ 64-128. In terms of parallelization efficiency, LLM inference on the H200 GPU outperforms the RTX-6000 Pro Blackwell GPU, and both are significantly more efficient than CPU-based parallelization (multi-processing) of the LexRank summarizer. We observe a very similar trend for SWE-Agent workload going from $BS=$ 64-128, where LLM inference parallelization on H200 is the most effective (1.06$\times$ increase in average latency), followed by LLM inference parallelization on RTX-6000 Pro Blackwell GPU (1.18$\times$ increase in average latency), followed by Bash multiprocessing on Intel GNR CPU ($1.53\times$ increase in average latency), followed by Bash multiprocessing on Grace CPU ($1.94\times$ increase in average latency).

\begin{takeaway}
    \textbf{Key Takeaway 3:} CPU-parallelization strategies fundamentally exhibit lower efficiency compared to GPU. In agentic AI workloads, they prematurely saturate the throughput, subsequently bottle-necking the system and degrading the utilization of costly GPU resources. 
\end{takeaway}

\section{Proposed Optimizations}

Based on throughput saturation insights (\autoref{subsection: throughput}), we present two scheduling optimizations- \circled{1} \textit{CPU-Aware Overlapped Micro-Batching} (COMB- \autoref{subsection:comb}) and \circled{2} \textit{Mixed Agentic Scheduling} (MAS- \autoref{subsection:mas}) for both homogeneous and heterogeneous agentic execution scenarios. We consider a practical serving scenario, namely the \textit{open-loop arrival system}.  In open-loop arrival system, requests are injected by an external arrival process, independent of the system state and prior completion information, thereby exposing the effects of queuing, resource contention, and scheduling decisions under sustained load.  We benchmark the performance of COMB and MAS under this system assumption to analyze their E2E performance efficacy. We measure request latency using percentile statistics: P50 denoting the 50th percentile or the median latency distribution; and P90 denoting the 90th percentile, i.e., the latency below which 90\% of requests get completed. Our objective is to reduce both P50 and P90, thereby improving not only median performance but also tail behavior, which is often the more critical metric in latency-sensitive serving systems.

\subsection{CPU-Aware Overlapped Micro-Batching (COMB)}
\label{subsection:comb}

On the CPU side, prior work \cite{garcia2022evaluating} shows that micro-batch granularity critically shapes the throughput–latency tradeoff in stream processing. In particular, micro-batch size and input frequency materially affect multi-core performance, with larger batches improving throughput only until CPU-side parallel efficiency degrades. LMStream \cite{lee2021lmstream} extends this idea to heterogeneous CPU–GPU streaming by dynamically controlling micro-batch admission to bound latency. In contrast, we focus on agentic pipelines that dynamically alternate between CPU-resident tool execution and GPU-resident model inference. Alternately, Ayo \cite{tan2025towards} adopts stage-local micro-batching, whereas we consider end-to-end micro-batching together with overlap across successive CPU and GPU stages. Finally, while \cite{recasens2025mind} introduces micro-batching for LLM inference, it targets GPU-centric execution. Our setting is different: we optimize CPU-induced micro-batches in agentic pipelines, whose preferred size is often smaller than GPU-induced LLM micro-batches. This is consistent with our observation of the fact that GPUs are fundamentally more efficient at parallelization than CPUs (refer \autoref{subsection: throughput}).

\begin{figure}[!t]
\centering
    \includegraphics[width=\columnwidth]
    {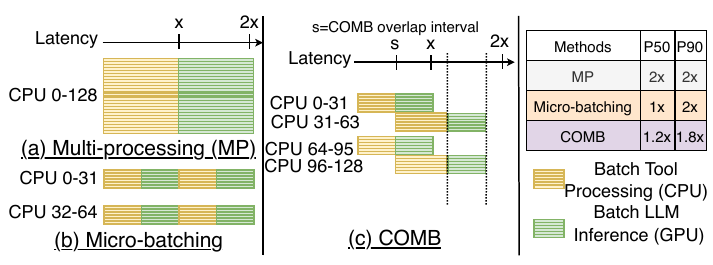}
    \caption{Timeline of batched agentic AI inference for (a) Multi-processing, (b) Microbatching, and (c) COMB.}
    \vspace{-3mm}
    \label{fig:comb_methodology}
\end{figure}

CPU-Aware Overlapped Micro-Batching (COMB) builds on prior work in micro-batching, but differs in both objective and mechanism. Specifically, rather than optimizing batching within an individual stage or device, COMB coordinates CPU-induced micro-batches across successive stages of an agentic pipeline to reduce inefficient CPU parallelization and temporal imbalance between CPU and GPU execution. As shown in \hyperref[fig:throughput]{Figure 3b}, CPU throughput saturates as $BS$ increases, and at $BS=$128, the median and tail latencies of the CPU-bound summarization stage both increase by $\sim2\times$. In addition, \hyperref[fig:comb_methodology]{Figure 4a} shows that CPU and GPU are heavily utilized in largely disjoint phases: \textbf{CPU-intensive tool execution leaves the GPU idle, while GPU-intensive inference leaves the CPU only lightly occupied for orchestration and runtime storage}. To tackle these inefficiencies, COMB first partitions a large incoming batch into a sequence of capped micro-batches of size at most $B_{cap}$. Based on empirical results, $B_{cap}\simeq1-2\times \#$ CPUs based on the parallelization efficiency of the specific CPU. This avoids over-subscription of CPUs and results in optimal CPU parallelism while preserving sufficient work to sustain GPU utilization. This design improves median and tail latency for large-batch execution by replacing the baseline’s monolithic batch of size $B_{max}$ ($B_{max} >$ $B_{cap}$), chosen to maximize GPU utilization. Micro-batching reduces CPU core oversubscription while yielding efficient CPU utilization. It also lowers instantaneous KV-cache demand, and preserves headroom for lightweight I/O-driven tools such as web search. In addition to micro-batch capping, \textit{COMB incorporates overlapping of adjacent micro-batches}, to mitigate device-level phase imbalance. As illustrated in \hyperref[fig:comb_methodology]{Figure 4c}, after an overlap interval $s$, once the CPU stage of micro-batch $i$ completes (e.g.: micro-batch 1 in \hyperref[fig:comb_methodology]{Figure 4c}, its GPU stage can execute concurrently with the CPU stage of micro-batch $i+1$ (e.g.: micro-batch 2 in \hyperref[fig:comb_methodology]{Figure 4c}. The result is a pipelined execution pattern that increases simultaneous CPU–GPU utilization, rather than optimizing micro-batch size alone.

\autoref{fig:comb_methodology} shows an example of COMB for $B_{cap} = 64$ and $B_{max}=128$ for a single-step agentic AI workload assuming throughput of individual stages (tool execution on CPU and LLM inference on GPU) saturate around $B_{cap}$. In case of micro-batching, the first micro-batch will finish around half of the total latency as the CPU contention is relieved while trading off the E2E tail latency. This is beneficial in cases of tiered serving system where different users are tiered differently based on amount of money they spend. Using COMB, the top 50\% of users can get $\sim2\times$ better service while maintaining the same service for the bottom 50\% tier of users compared to the baseline. The overlapping trades-off utilization for some of the P50 gains observed during micro-batching. Micro-batching is highly effective when we observe complete throughput saturation going from $B_{cap}$ to $2 \times B_{cap}$. However, overlapped micro-batching helps in optimizing more general agentic workloads which can have partial throughput saturation profile due to concurrent CPU-GPU utilization. 
For such workloads, compared to MP, COMB trades a modest increase in tail latency for substantial reductions in P50 latency. Additionally, varying the overlap duration $s$ can yield a P50–P90 latency Pareto frontier.

\subsubsection{Throughput Gain and COMB Effectiveness}

Let $T(BS)$ denote the throughput ($\frac{BS}{t_{sec}}$). In our example, we define the throughput gain ratio as $r(BS) = \frac{T(BS = 2^n)}{T(\frac{BS}{2} = 2^{n-1})}$ which captures the speedup achieved by halving the batch size. Here, $n$ represents the batch multiplicative factor. \autoref{tab:throughput_gain_ratio} shows the value of $r(64)$ and $r(128)$ for Web-Augmented Agent and SWE-agent on the two hardware platforms. For micro-batching, the optimal $B_{cap}$ should maximize resource efficiency while avoiding the saturation regime where additional parallelism yields negligible improvements. If the gain ratio $r(BS) \simeq 1$, micro-batching will be highly effective and save $\sim2\times$ P50 latency while preserving similar P90 latency. On the other hand, for $r(BS) > 1.5$, there is little or no throughput saturation and micro-batching will be ineffective. If  $1 < r(BS) < 1.5$, overlapped micro-batching with different overlap durations will result in a P50-P90 Pareto frontier and will be more effective than just micro-batching due to better CPU-GPU utilization. \textbf{The effectiveness of COMB is inversely proportional to the gain ratio as we will see later during the empirical evaluation}.

\begin{table}[]
\centering
\small
\caption{Throughput gain ratios for Web-Augmented Agent and SWE-Agent workloads on both the systems.}
\label{tab:throughput_gain_ratio}
\begin{tabular}{l|ll|ll}
\hline
\multirow{2}{*}{Workload}                                       & \multicolumn{2}{l}{\texttt{Sys 1}}          & \multicolumn{2}{l}{\texttt{Sys 2}}          \\ \cline{2-5} 
                                                                & \multicolumn{1}{l|}{r(64)} &128$r(128)$ & \multicolumn{1}{l|}{$r(64)$} & $r(128)$ \\ \hline
\begin{tabular}[c]{@{}l@{}}Web-Augmented  \\ Agent (LangChain)\end{tabular} & \multicolumn{1}{l|}{1.94}     & 1.00      & \multicolumn{1}{l|}{1.76}     & 1.05      \\ \hline
SWE-Agent                                                       & \multicolumn{1}{l|}{1.43}     & 1.18      & \multicolumn{1}{l|}{1.45}     & 1.15      \\ \hline
\end{tabular}
\end{table}

\subsubsection{COMB in Open-Loop Agentic Serving Systems}
In open-loop arrival systems, we do not impose explicit overlapping due to probabilistic arrivals and variation in request service times. Instead, we employ concurrency cap ($N_{cap}$) as a practical online approximation to the COMB principle. By limiting concurrency of CPU-heavy work, the scheduler naturally interleaves CPU-dominant and GPU-dominant phases across requests without rigid overlap. This prevents over (or under)-utilization of one of the CPU or GPU resources compared to the server-induced higher concurrency cap ($N_{max}$) and ensures better CPU-GPU concurrent utilization. Let us assume, user request arrival rate to be $\lambda$ with mean service time across requests being $\mathbb{E}[\mathcal{S}]$  under $m$ number of hardware resources. We define the utilization $\rho$ as, $\rho = \frac{\lambda \mathbb{E}[\mathcal{S}]}{m}$. If we derive $N_{max}$ to extract optimum performance out of the costly GPU resource in a datacenter server, the GPU utilization is close to 1 ($\rho_{GPU}=1$). This in turn results in a large $N_{max}$ that creates over-utilization of CPU ($\rho_{CPU} > 1.5$) in agentic workloads having significant tool execution stage. Although, the high concurrency cap was chosen to extract optimum GPU utilization, the CPU over-utilization starves the GPU and results in GPU under-utilization. In utilization terminology, COMB-induced $N_{cap}$ reduces $\rho_{CPU}$ and increases $\rho_{GPU}$, thereby balancing CPU-GPU utilization.

\subsection{Mixed Agentic Scheduling (MAS)}
\label{subsection:mas}

As discussed in \autoref{subsection: latency profiling}, agentic workloads are inherently heterogeneous: some requests are \emph{CPU-heavy}, dominated by tool execution on the host, while others are \emph{GPU-heavy}, purely LLM inference on the GPU. COMB optimization targets the CPU-heavy regime, where the critical path contains substantial host-side tool execution. However, many practical deployments must also serve GPU-heavy requests with no tool use. For example, even within a single chatbot service, some requests invoke external tools while others are handled largely by direct LLM inference. 

Prior serving systems such as vLLM \cite{kwon2023efficient} and SGLang \cite{zheng2024sglang} optimize homogeneous LLM inference, where scheduling is driven primarily by GPU throughput. %
For example, \textsc{vLLM} increases throughput through paged attention and continuous batching. These designs are highly effective when requests contend for essentially the same hardware resource. They are less well matched to heterogeneous agentic serving, where CPU-heavy and GPU-heavy requests stress different bottlenecks and can interfere with one another if admitted through a single queue. In such settings, a request mix skewed toward one request-type can monopolize admission, causing the other request-type to experience inflated wait time despite having a different resource bottleneck.

Mixed Agentic Scheduler (MAS) is built around two complementary policies. First, it performs \emph{request-type-aware concurrent admission} for CPU- and GPU-heavy requests using separate execution queue caps, $E_{\text{cap,CPU}}$ and $E_{\text{cap,GPU}}$, respectively for CPU-heavy requests and GPU-heavy requests. This policy allows the system to exploit both resource domains at the same time instead of serializing them through a single queue. Concretely, CPU-heavy requests are admitted to an execution queue bounded by $E_{\text{cap, CPU}}$, while GPU-heavy requests are admitted to a separate execution queue bounded by $E_{\text{cap, GPU}}$. Requests that exceed these request-type specific caps are placed into a \emph{shared reserved execution queue} of size $E_{\text{cap, shared}}$ providing the elasticity in concurrency of either of the request-type beyond the request-specific execution queue caps. On the CPU side, we derive $E_{\text{cap, CPU}}=N_{cap}$ from COMB evaluation in the open-loop arrival settings to improve host-side concurrency for CPU-heavy requests. On the GPU side, we allocate the remaining concurrency budget for $E_{\text{cap, GPU}}$ out of $N_{max}$ as large as possible for effective GPU utilization. These elastic caps for requests preserve work conservation while preventing one request-type from fully monopolizing admission under asymmetric (one request-type dominate the other) open-loop arrivals. For example, the dominant request-type will occupy most of the concurrency slots out of $N_{max}$ in the baseline and the minority request-type will suffer long queuing delays due to limited concurrency available. MAS protects the minority request-type by allotting a minimum concurrency of $E_{\text{cap, CPU}}$ for CPU-heavy request type or $E_{\text{cap, GPU}}$ for GPU-heavy request type (whichever is the minority request-type). Together, these two policies allow MAS to reduce cross-request interference, sustain concurrent CPU-GPU utilization, and improve performance for heterogeneous open-loop arrivals. The MAS algorithm is detailed in \autoref{appendix:mas_algorithm}.

\vspace{-3mm}
\section{Experimental Evaluations}
We first include a single-batch experiment to illustrate how COMB improves concurrent CPU-GPU utilization without inter-request interference. In this standalone setting, we also visualize the P50–P90 Pareto frontier to characterize the tradeoff between median and tail latency. We then evaluate our optimizations in the open-loop arrival setting and sweep arrival rates to study performance under sustained load. We then present an ablation study on a resource-constrained system having limited CPU cores to demonstrate the generality of the proposed optimizations under tighter host-side bottlenecks. Additionally, we present a detailed energy profiling that reveals the substantial dynamic energy overhead of CPUs in CPU-centric agentic AI.

\subsection{COMB: Analysis}

\begin{figure*}
    \centering
    \includegraphics[width=1\linewidth]{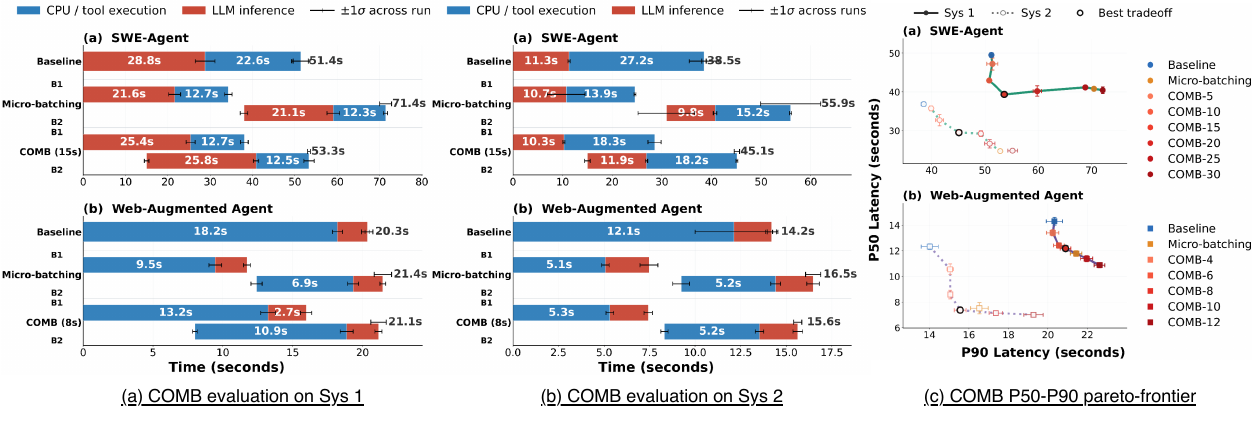}
    \caption{COMB evaluation for standalone batch processing of $BS$=128 showing better CPU-GPU utilization for Web-Augmented Agent and SWE-Agent workloads on (a) \texttt{Sys 1} and (b) \texttt{Sys 2} with (c) P50-P90 Pareto frontier for different overlap intervals.}
    \label{fig:COMB_eval}
    \vspace{-1em}
\end{figure*}

In \autoref{fig:COMB_eval}, we conduct an experiment to evaluate COMB in a standalone concurrent processing with $BS = 128$ requests. The baseline strategy is to perform multi-processing on all the 128 requests, while COMB uses a $B_{cap}$ of 64 derived from the throughput gain ratio analysis in \autoref{tab:throughput_gain_ratio}. We observe that the P50/P90 gains (\hyperref[fig:COMB_eval]{Figure 5c}) from COMB is inversely proportional to the throughput gain ratio, $r(BS)$. For SWE-Agent workload, $r(128)= 1.15$ and $1.18$ for \texttt{Sys 1} and  \texttt{Sys 2}, respectively, while $r(128)\simeq1$ for Web-Augmented Agent on both the systems. As a result, micro-batching is highly effective for Web-Augmented Agent workload resulting in 1.65$\times$ speedup in P50 latency while slowing down the P90 latency by a factor of 0.86$\times$ on \texttt{Sys 2}. On the other hand, micro-batching is not effective for SWE-Agent resulting in slow-down by a factor of 0.72$\times$ and 0.69$\times$ in P90 latencies on \texttt{Sys 1} and \texttt{Sys 2}, respectively, significantly worsening the tail performance.
In \hyperref[fig:COMB_eval]{Figure 5c}, we further plot the Pareto frontier for different overlap values, $s$ for COMB. We choose the optimal $s$ to represent the best P50-P90 trade-off close to the knee of the Pareto frontier. For example, with Web-Augmented Agent on \texttt{Sys 2}, we observe that at $s=8s$, COMB yields moderately accelerated latency compared to micro-batching by  1.03$\times$ and 1.05$\times$ for P50 and P90, respectively. On the other hand, for SWE-Agent on \texttt{Sys 1}, we observe that for $s=15s$, COMB yields 1.40$\times$ improved P90 latency while achieving similar P50 latency compared to micro-batching.

\subsection{Open-Loop COMB: Analysis}
\label{subsec: open-loop COMB}

\begin{figure}
    \centering
    \includegraphics[width=0.92\linewidth]{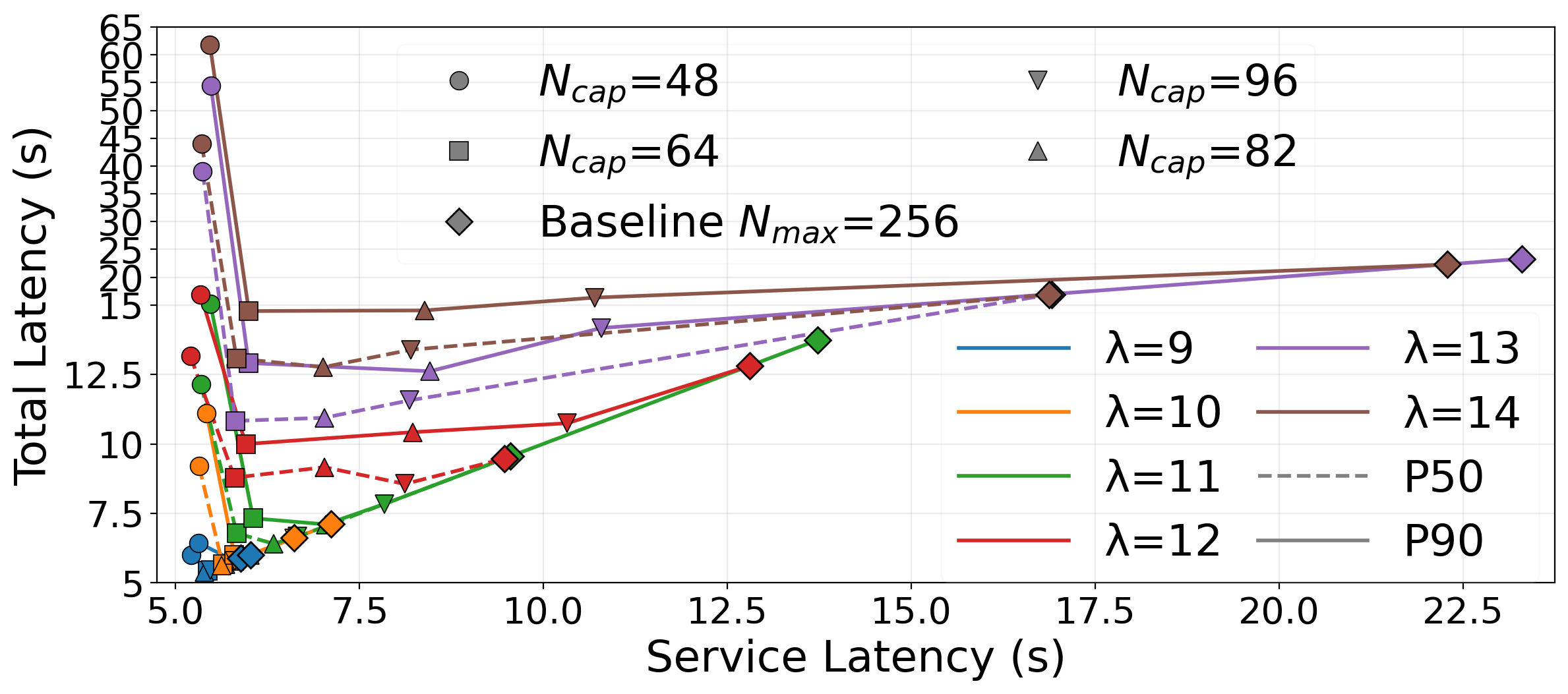}
    \caption{In open-loop serving system with $\lambda=9-15$ req/s arrival-rate for Web-Augmented Agent on \texttt{Sys 2}.}
    \label{fig:COMB_eval_open_loop_system2}
    \vspace{-1em}
\end{figure}

For the baseline, we set the concurrency cap ($N_{max}$) to maximize GPU utilization while avoiding diminishing returns. Thus, $N_{max}$ is chosen to be the knee of the throughput-batch size curve where the increase in throughput saturate under a $2\times$ increase in batch size. As shown in \autoref{tab:gpu_saturation_table}, this condition is met at batch size 256 on both RTX-6000 Pro Blackwell and H200 GPU for the LLM inference stage of the LangChain workload. Accordingly, we set $N_{max}=256$ for the baseline. 
COMB is evaluated under the open-loop arrival setting as shown in \autoref{fig:COMB_eval_open_loop_system2} on \texttt{Sys 2}. As shown in the figure, the baseline becomes increasingly CPU-overloaded as the Poisson arrival rate increases, with $\rho_{CPU}=1.54, 1.66, 3.09,$ and $3.18$ for $\lambda=11,12,13,$ and $14$ req/s, respectively. In contrast, for COMB with $N_{cap}=64$, $\rho_{CPU}$ remains in the narrower range of $0.89-1.13$ over $\lambda=11-14$ req/s, yielding the best service latency among other concurrency cap configurations. The strongest gains over the baseline appear at higher loads. At $\lambda=13$ req/s, COMB reduces service latency by 2.9$\times$ and 3.9$\times$ at the P50 and P90 percentiles, respectively. On the other hand, total latency\footnote{Here, total latency is computed by averaging the net latency over all the requests during a specific period, with net latency for each serving request is its service latency + wait latency.} also drops by 1.6$\times$ and 1.8$\times$ at the P50 and P90 percentiles, respectively, compared to the baseline. Consequently, the COMB-induced $N_{cap}=64$ results in a throughput improvement of 1.7$\times$ compared to the baseline. %
More aggressive concurrency capping with ($N_{cap}=48$) lowers the service time, however, sharply increases queuing delay. On the other hand, larger $N_{cap}$ of 82 and 96 provide comparable throughput with worse service latency than that with $N_{cap}=64$. This supports the choice of $N_{cap}=64$ in yielding a balance between utilization and queuing delay.

\begin{table}[!t]
\centering
\small
\caption{vLLM TPS (Token/s) for LLM part of LangChain workload on both systems shows that B=256 gives the best GPU utilization at the knee of the throughput-batch  curve.}
\label{tab:gpu_saturation_table}
\begin{tabular}{l|l|l|l|l|l}
\hline
$BS$ & \begin{tabular}[c]{@{}c@{}} \texttt{Sys 1} \\ TPS \end{tabular} & \begin{tabular}[c]{@{}c@{}} \texttt{Sys 2} \\ TPS \end{tabular} & $BS$  & \begin{tabular}[c]{@{}c@{}} \texttt{Sys 1} \\ TPS \end{tabular} & \begin{tabular}[c]{@{}c@{}} \texttt{Sys 2} \\ TPS \end{tabular} \\
\hline
1     &   1313.27  & 1566.13    &  32 & 12500.29 & 13846.48 \\ 
\hline
2     & 2089.84 & 2547.38 & 64    & 15264.86 & 19664.15 \\
\hline
4     & 3950.28 & 4615.84  & 128    & 17326.58 & 23504.65 \\ 
\hline
8     & 6458.83 & 7051.94 & 256   & 18779.70 & 29138.36 \\
\hline
16 & 8960.46 & 9846.26 & 512 & 19468.49 & 32860.12 \\
\hline
\end{tabular}
\end{table}

\subsection{MAS: Analysis}

\begin{figure}[t]
    \centering

    \begin{subfigure}{\linewidth}
        \centering
        \includegraphics[width=0.95\linewidth]{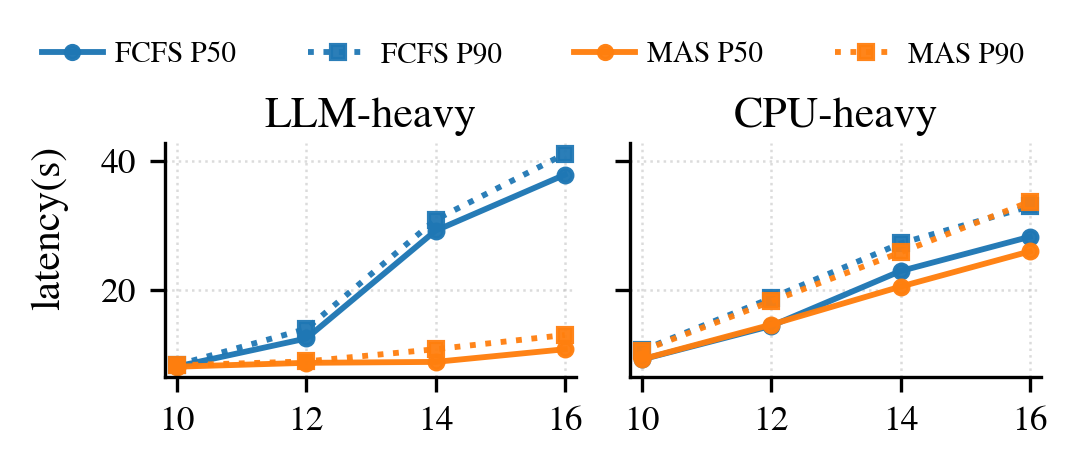}
        \caption{Request type probability, $p_{LLM}$ = 0.25}
        \label{fig:mas_eval_system2_prob_25}
    \end{subfigure}

    \begin{subfigure}{\linewidth}
        \centering
        \includegraphics[width=0.95\linewidth]{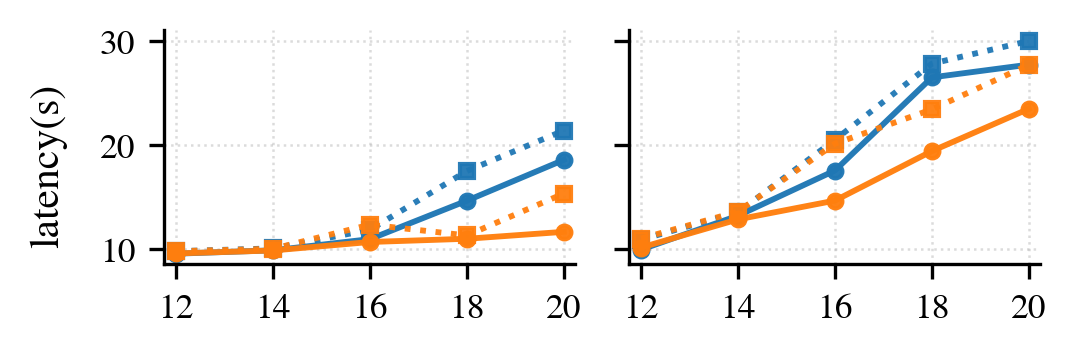}
        \caption{Request type probability, $p_{LLM}$ = 0.50}
        \label{fig:mas_eval_system2_prob_50}
    \end{subfigure}

    \begin{subfigure}{\linewidth}
        \centering
        \includegraphics[width=0.95\linewidth]{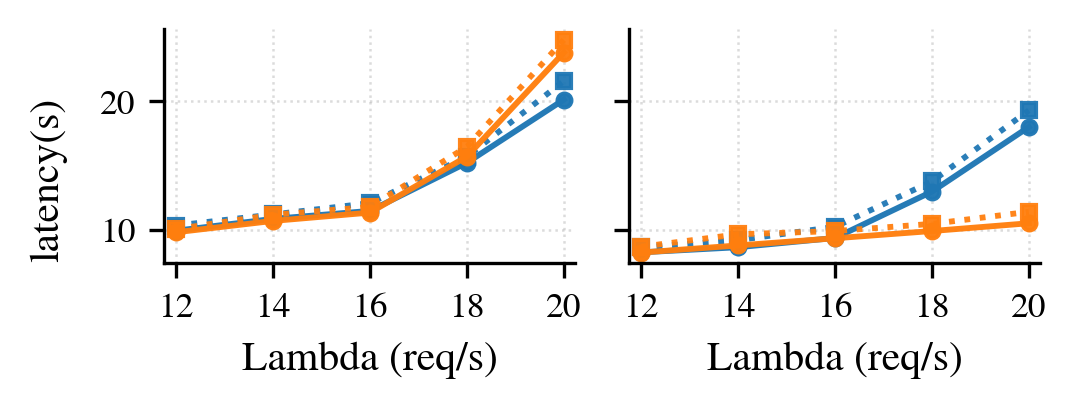}
        \caption{Request type probability, $p_{LLM}$ = 0.75}
        \label{fig:mas_eval_system2_prob_75}
    \end{subfigure}
        \vspace{-5mm}
        \caption{Iso-concurrency evaluation of MAS relative to the FCFS baseline for heterogenous requests under bursty arrival patterns for different $p_{LLM}$ on \texttt{Sys 1}.}
        \vspace{-7mm}
    \label{fig:mas_eval_system1}
\end{figure}

\begin{figure}[t]
    \centering

    \begin{subfigure}{\linewidth}
        \centering
        \includegraphics[width=0.95\linewidth]{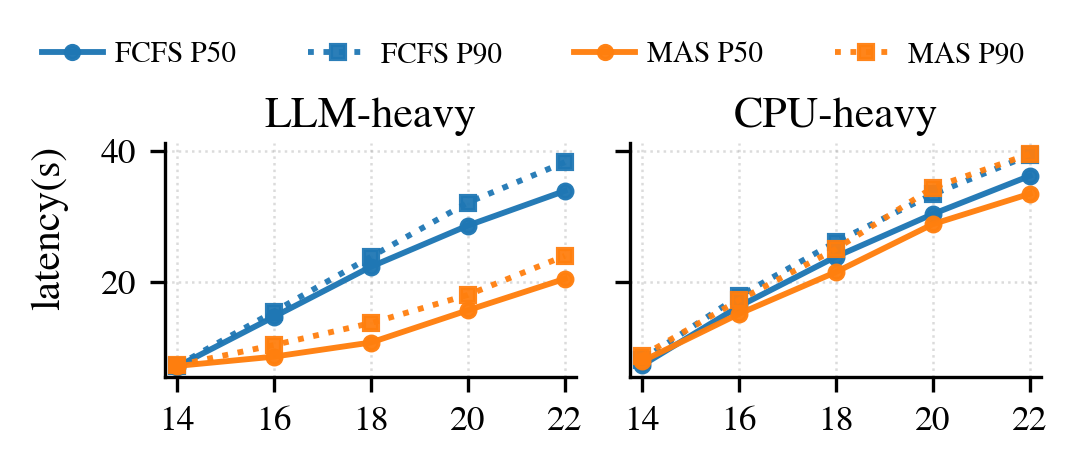}
        \caption{Request type probability, $p_{LLM}$ = 0.25}
        \label{fig:mas_eval_system2_prob_25}
    \end{subfigure}

    \begin{subfigure}{\linewidth}
        \centering
        \includegraphics[width=0.95\linewidth]{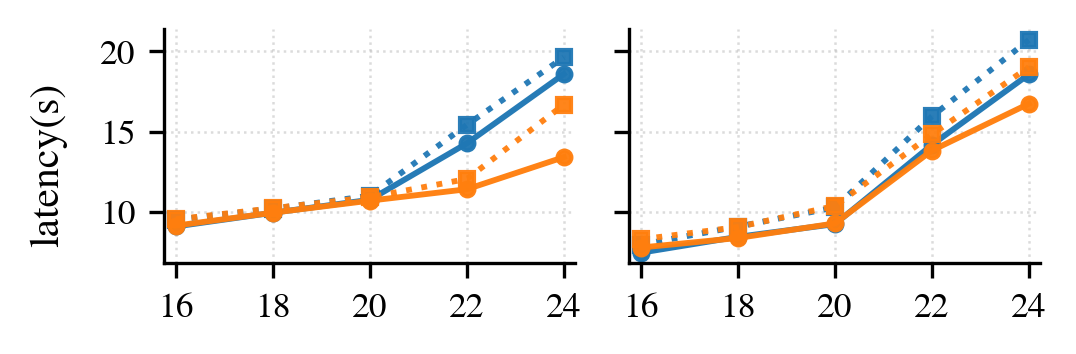}
        \caption{Request type probability, $p_{LLM}$ = 0.50}
        \label{fig:mas_eval_system2_prob_50}
    \end{subfigure}

    \begin{subfigure}{\linewidth}
        \centering
        \includegraphics[width=0.95\linewidth]{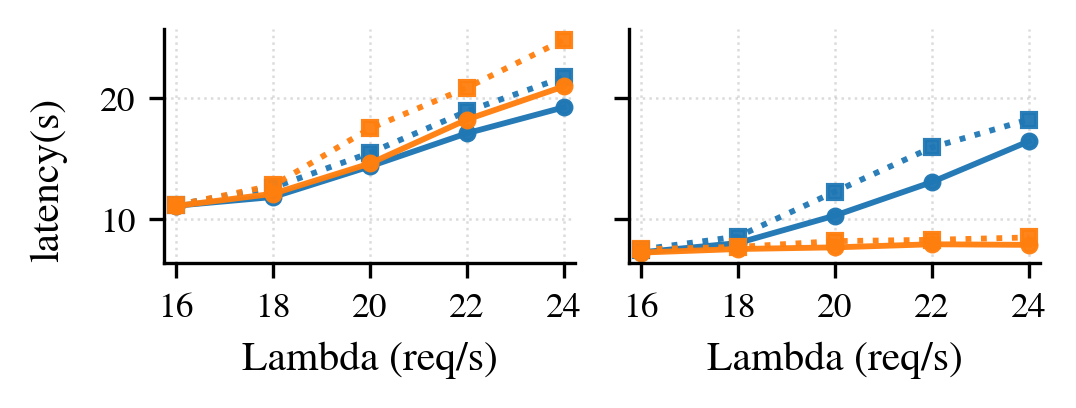}
        \caption{Request type probability, $p_{LLM}$ = 0.75}
        \label{fig:mas_eval_system2_prob_75}
    \end{subfigure}
    \vspace{-5mm}
        \caption{Iso-concurrency evaluation of MAS relative to the FCFS baseline for heterogenous requests under bursty arrival patterns for different $p_{LLM}$ on \texttt{Sys 2}.}
    \vspace{-5mm}
    \label{fig:mas_eval_system2}
\end{figure}

On both \texttt{Sys 1} and \texttt{Sys 2}, we set a common iso-concurrency budget of $N_{\max}=224$ for both FCFS and MAS. We chose this value of $N_{\max}$ as it is close to the knee of GPU saturation for both the systems (refer to \autoref{tab:gpu_saturation_table}). For MAS, we partition this budget into a CPU-heavy admission cap of $E_{\text{cap,CPU}}=N_{cap}=64$, derived from the open-loop COMB study (\autoref{subsec: open-loop COMB}), a shared reserved queue of size $E_{\text{cap,shared}}=32$, and the remaining budget is allocated to GPU-heavy requests, yielding $E_{\text{cap,GPU}}=128$. We evaluate MAS under three request mixes of GPU-heavy request arrival probabilities $p_{\mathrm{LLM}}$. We choose  $p_{\mathrm{LLM}} \in \{0.25, 0.50, 0.75\}$, each value denoting the probability of an arriving request being GPU-heavy (i.e., pure LLM inference), with $1-p_{\mathrm{LLM}}$ denoting probability of a request being CPU-heavy. In MAS analysis of \autoref{fig:mas_eval_system1} and \autoref{fig:mas_eval_system2}, we plot the total latency of each request type during a steady-state period of 400 requests over a total of 1500 requests. If steady-state is not reached, we consider 400 requests at the center across the total 1500 requests. To stress-test scheduling under bursty load, we drive the system with a Poisson arrival process of rate $\lambda$ whose request-type mix follows a two-state ON/OFF model \cite{heffes1986markov}, where ON/OFF phase changes after every 32 requests.

MAS improves fairness by aligning admission with each request type bottleneck: the CPU-heavy elastic cap, $E_{\text{cap,CPU}}=64$ avoids oversubscribing host cores, the GPU-heavy cap keeps the GPU well utilized, and the reserved queue limits burst-induced head-of-line blocking. This request-aware admission policy consistently protects the minority request under skewed mixes on both systems. When $p_{\mathrm{LLM}}=0.25$, both systems primarily benefit the minority GPU-heavy requests. On \texttt{Sys 2}, MAS improves GPU-heavy latency by up to $1.82\times/1.78\times$ at $P50/P90$ percentiles. On \texttt{Sys 1}, the improvement is even larger, reaching up to $2.37\times/2.49\times$, with CPU-heavy request total latency remaining largely unchanged. At $p_{\mathrm{LLM}}=0.50$, the gains become more balanced on \texttt{Sys 2}, reaching $1.39\times/1.18\times$ for GPU-heavy $P50/P90$ latencies, with roughly $1.1\times$ improvement for CPU-heavy requests. When $p_{\mathrm{LLM}}=0.75$, MAS instead protects the minority CPU-heavy requests against an LLM-dominated arrival load. On \texttt{Sys 2}, it keeps GPU-heavy latency nearly flat across the sweep and yields $2.09\times$ and $2.15\times$ improvements in CPU-heavy $P50$ and $P90$ latencies, respectively, at $\lambda=24$. \textbf{In terms of total latency benefit, for instance, in \texttt{Sys 1} at $p_{LLM}=0.50$,  MAS yields an average (over all requests) speed up of 1.62$\times$/1.30$\times$ in P50/P90 latency}. Overall, \textit{MAS prevents the dominant request-type from monopolizing the systems, improving both total latency and the fairness of CPU and GPU utilization under heterogeneous open-loop load}.

\subsection{Ablation Studies}

\subsubsection{Ablation on a CPU-Core Constrained Platform}
To evaluate the effectiveness of our optimizations beyond \texttt{Sys 1} and \texttt{Sys 2}, we perform ablation on a third platform consisting of a 16-core Intel Emerald Rapids CPU paired with the same RTX-6000 Pro Blackwell GPU (similar to in \texttt{Sys 1}). Relative to the 64-core and 72-core hosts in the first two systems, this platform provides roughly one quarter of the CPU capacity while keeping the accelerator unchanged, thereby isolating a substantially tighter host-side bottleneck.

\begin{figure}
    \centering
    \includegraphics[width=0.75\linewidth]{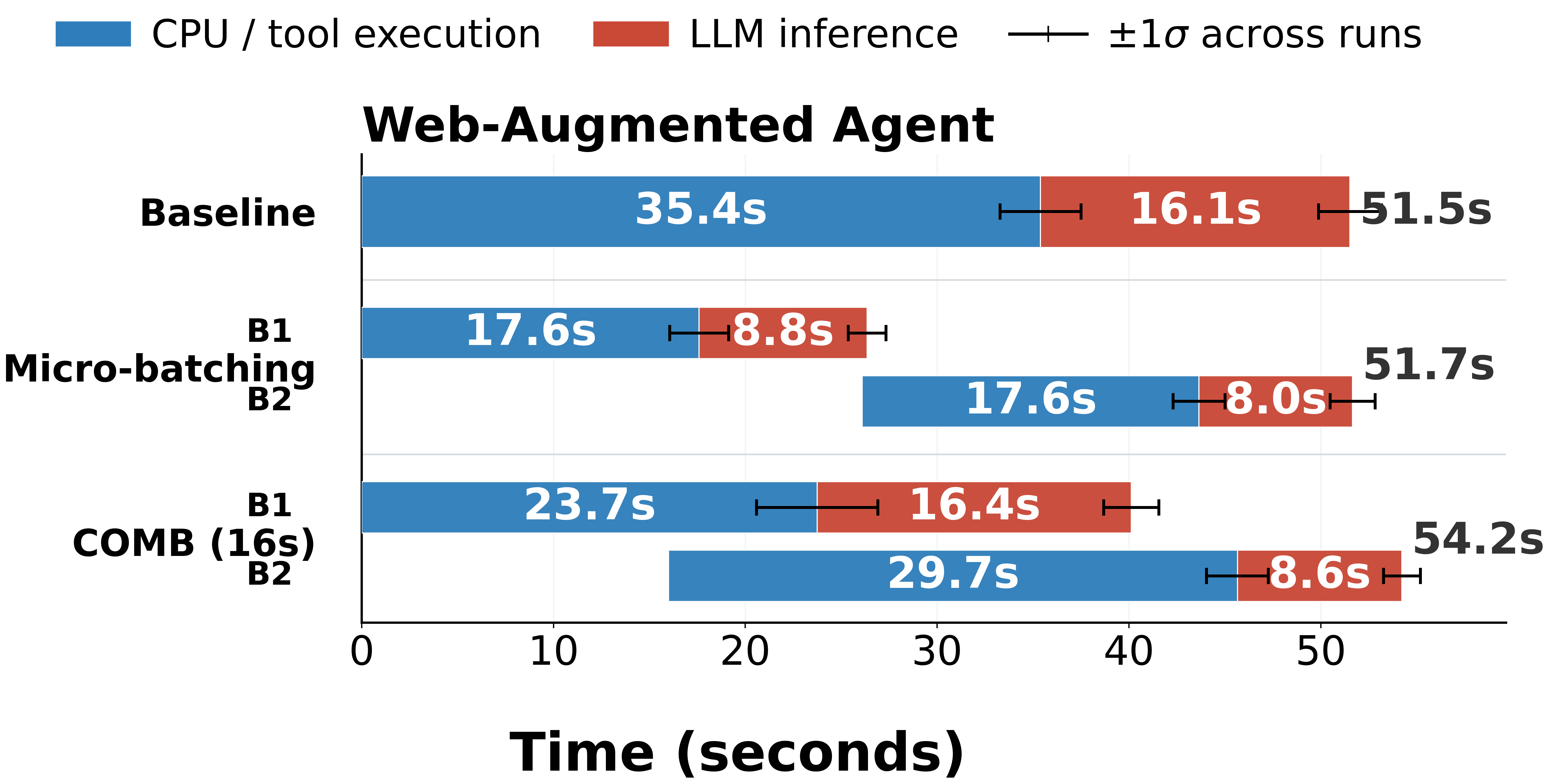}
    \vspace{-4mm}
    \caption{Ablation: COMB evaluation on 16-core CPU system for Web-Augmented Agent.}
    \label{fig:comb_ablation}
    \vspace{-6mm}
\end{figure}

\noindent
\textbf{COMB.} In \autoref{fig:comb_ablation}, we present results on the CPU- constrained system. We use the same Web-Augmented Agent workload and standalone batch processing setup with $B_{max}=64$ for baseline and $B_{cap}=32$ for COMB. We observe that micro-batching is highly effective as the gain ratio $r(64)\simeq1$, however, the benefit of overlap becomes more sensitive to CPU availability. On this system, micro-batching reduces the first-batch completion time from 51.5s to 26.4s, yielding a $1.95\times$ improvement, while leaving the tail latency nearly the same at 51.7s. In contrast, COMB with overlap duration $s=$16s increases the first-batch completion time to 40.1s due to higher CPU-side contention. Overall, \textit{this shows relative ineffectiveness of COMB on a CPU-core limited system, as micro-batching alone can improve P50 and P90 latency by 1.52$\times$ and 1.05$\times$, respectively when the gain ratio $r(BS)\simeq1$}.

\begin{figure}
    \centering
    \includegraphics[width=0.9\linewidth]{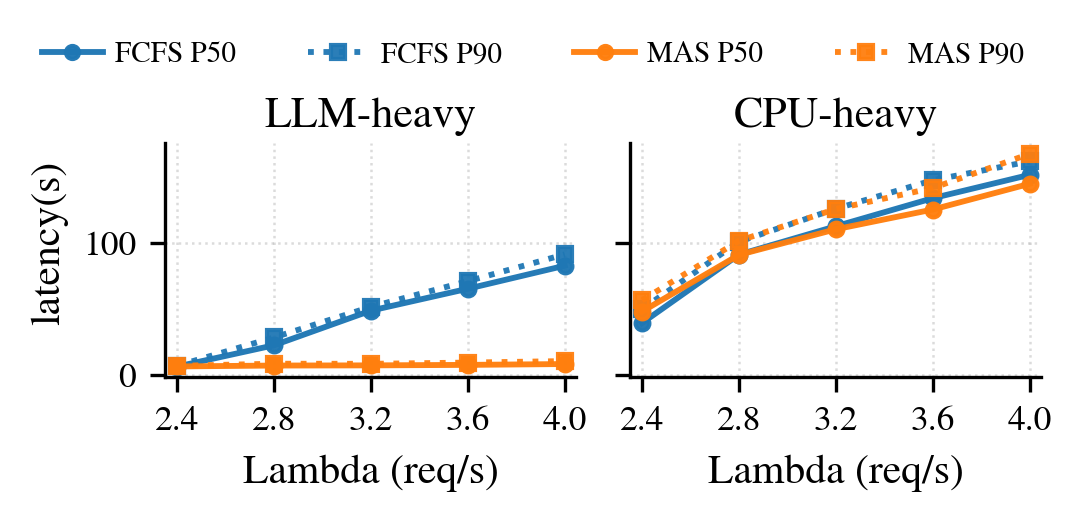}
    \vspace{-5mm}    
    \caption{Ablation: MAS evaluation on 16-core CPU for GPU-heavy request-type probability, $p_{LLM}$=0.25}
    \label{fig:mas_ablation}
    \vspace{-5mm}
\end{figure}

\noindent
\textbf{MAS.} For this setting, we retain the same admission cap for GPU-heavy requests used for \texttt{Sys 1} (as it has the same GPU), but reduce the CPU-heavy cap to $E_{\text{cap,CPU}}=32$ based on empirical evaluation on the third platform. This setup tests whether the same request-aware admission principle continues to hold under a much more CPU-constrained regime. \autoref{fig:mas_ablation} shows that under the representative skewed mix with $p_{\mathrm{LLM}}=0.25$, MAS continues to protect the GPU-heavy minority request without sacrificing throughput, despite the much smaller CPU budget. At light load, FCFS and MAS behave similarly, but as load increases, the benefit of separating CPU-heavy and GPU-heavy admissions becomes pronounced. In particular, GPU-heavy request total latency improves by up to $10.1\times$ at $P50$ and $8.8\times$ at $P90$, while throughput remains essentially unchanged at low load and improves by up to $1.06\times$ near the high-load end of the sweep. At the same time, CPU-heavy latency changes only modestly, indicating that these gains do not come from starving the CPU-heavy majority, but from preventing it from monopolizing the total concurrency budget. Moreover, MAS results in $\sim1.20\times$ P50/P90 speedup across all the requests. Overall, this ablation strengthens the main conclusion of our design: MAS is not tied to the larger CPU budgets of \texttt{Sys 1} and \texttt{Sys 2}, but generalizes to systems with substantially smaller CPU-GPU ratios, where request-type-aware admission becomes even more important for preserving latency isolation while maintaining throughput.

\subsubsection{Ablation on Dynamic CPU-GPU Energy Profiling.}
\label{ablation:energy_profiling}
For \texttt{Sys 2}, both CPU and GPU energy were obtained via \textit{nvidia-smi}. \textit{Nvidia-smi} reads module power and we get CPU energy by subtracting GPU power from module power. In the quiescent (idle) state, for \texttt{Sys 2}, the Grace CPU drew 140 W while the H200 GPU drew 142 W. We run each of the workloads five times to account for statistical variance.

\begin{figure}
    \centering
    \includegraphics[width=1\linewidth]{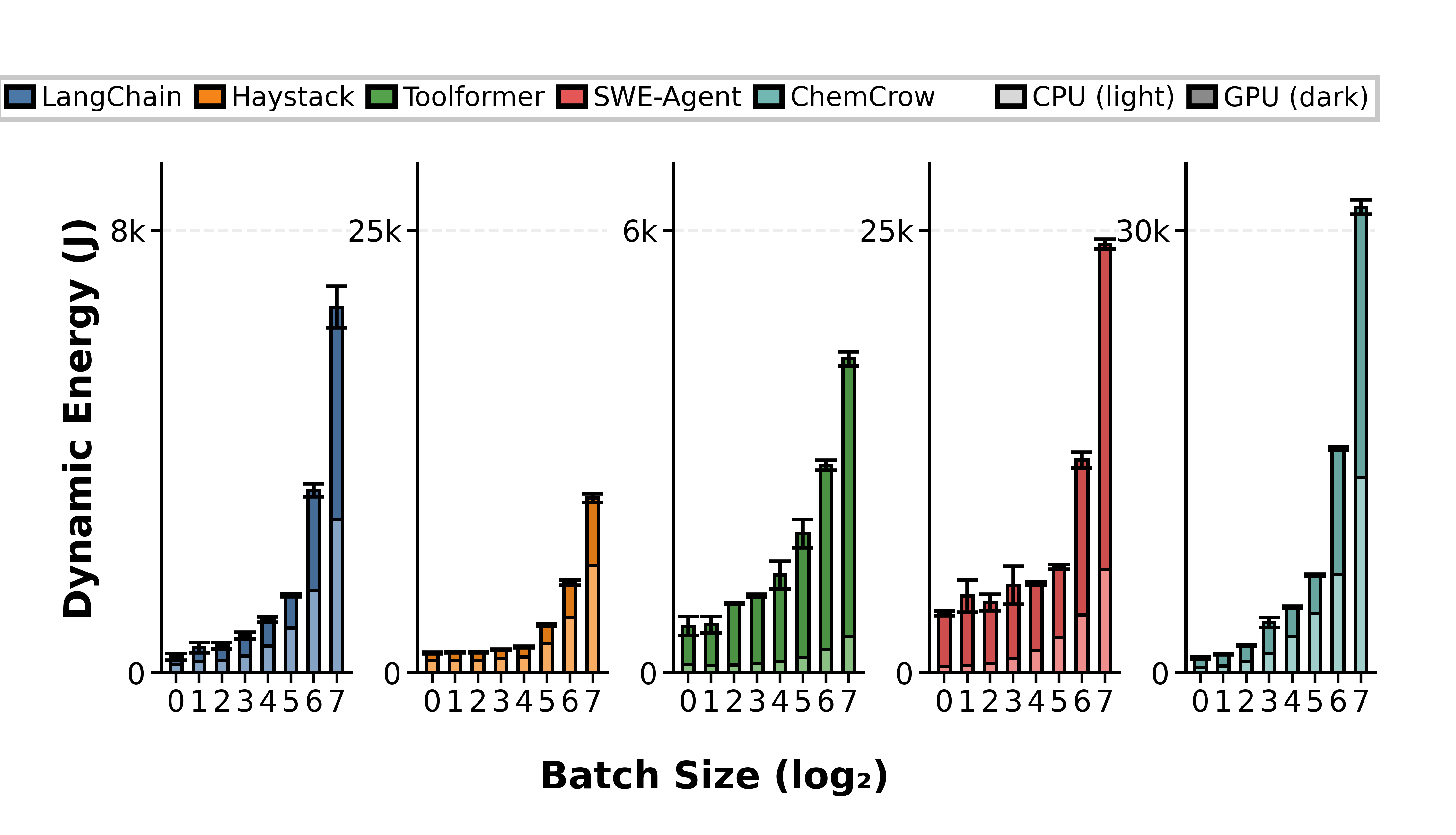}
    \vspace{-5mm}
    \caption{CPU and GPU dynamic energy consumption.}
    \label{fig:energy}
    \vspace{-5mm}
\end{figure}

\autoref{fig:energy} shows that across all batch sizes, RAG (Haystack) exhibits a uniform CPU dynamic energy contribution of 61\% to total system dynamic energy. For Web-Augmented Agent (LangChain), CPU accounts for 43-52\% of total dynamic energy for small $BS=1-8$. As $BS$ increases to 16 and 32, CPU dynamic energy share rises to 50\% and 57\%. However, CPU dynamic energy share decreases to 42\% at $BS=128$. ChemCrow shows a similar rising-then-falling pattern, with CPU contributing 35-40\% of total dynamic energy for small $BS=1-8$, climbing to 55\% ($BS = 16$) and peaking at 60\% ($BS=32$), before dropping to 42\% ($BS=128$). On the other hand, Toolformer workload exhibits a consistently low and slightly decreasing CPU dynamic energy utilization as the batch size increases. CPU contributes 11-18\% dynamic energy at $BS=1-8$ and the contribution settles to 11-12 at large batch sizes since the workload is primarily GPU-heavy. For SWE-agent workload, CPU contributes only 10-16\% to total dynamic energy at lower batch sizes $BS=1-8$, but the contribution increases sharply as the batch size goes up with 25\% at $BS=64$ and 33\% at $BS=128$ respectively. \textit{CPU dynamic energy share becomes significant (up to 61\%) for CPU-centric agentic AI workloads, motivating CPU-centric energy optimization policies in future agentic AI data-centers}.

\section{Related Works}
\textbf{Agentic AI Characterization.} A recent work \cite{sapkota2025ai} characterized agentic AI based on agentic capabilities. In contrast, we characterize agentic AI from compile-time algorithmic point of view. Another work \cite{{vellaisamy2025characterizingoptimizingllminference}} performed runtime characterization based on kernel launch delay while our runtime characterization is based on E2E latency and throughput.

\noindent
\textbf{Agentic AI Profiling.} 
A recent work \cite{kim2025cost} profiled agentic workloads from a GPU-centric perspective without exposing the CPU bottleneck due to tool processing. Most of the tools they used are API calls (WolframAlpha and Wikipedia) and can easily be parallelized. Another work \cite{asgar2025efficient} profiled agentic AI workloads and optimized the orchestration framework but focused solely on external tool calls. Therefore, the work was based on nearly zero local CPU overhead, lacking a comprehensive CPU-centric perspective. 

\noindent
\textbf{Scheduling Optimizations.} Prior work has studied micro-batching from several complementary perspectives. On the CPU side, LMStream \cite{lee2021lmstream} and \cite{garcia2022evaluating} show that micro-batch granularity affects the throughput-latency tradeoff in streaming systems. Ayo \cite{tan2025towards} adopts stage-local micro-batching. In contrast, COMB introduces end-to-end micro-batching for agentic workloads, coordinating CPU-induced micro-batches across successive CPU and GPU stages to reduce inefficient CPU parallelization and cross-device imbalance. vLLM \cite{kwon2023efficient} and SGLang \cite{zheng2024sglang} optimize scheduling for GPU-only serving systems. Our approach orthogonally use continuous batching for LLM inference like vLLM and SGLang with elastic queuing for heterogeneous agentic requests along with a reserved queue to sustain bursty traffic.

\section{Conclusions}

Agentic AI shifts the system bottleneck from monolithic LLM inference toward CPU-resident tool execution and orchestration. In this work, we characterize representative agentic workloads from a CPU-centric perspective and show that these workloads exhibit CPU latency and throughput bottlenecks. To tackle these bottlenecks, we introduce \textit{COMB} and \textit{MAS}, two scheduling techniques for homogeneous and heterogeneous agentic workloads, respectively. Together, these optimizations yield improved CPU-GPU concurrent utilization while reducing skewed resource allocation for heterogeneous execution.

\begin{acks}
We thank Sarbartha Banerjee, Zishen Wan, Akshat Ramachandran and Shubham Jain for thoughtful discussions and  valuable feedback that helped improve this work. 
\end{acks}

\bibliographystyle{ACM-Reference-Format}
\bibliography{ref}

\appendix

\section{Representative Workloads}
\label{appendix: representative workloads}
\subsection{Toolformer}
Toolformer teaches language models to use external tools through self-supervised learning~\cite{schick2023toolformer}. It teaches the GPT-J 6B \cite{gpt-j} model to decide when and how to call tools like calculators, QA systems, and search engines. It achieves 40.4\% accuracy on ASDiv math problems \cite{miao2021diverse}, outperforming the GPT-3 175B model.

\subsection{SWE-Agent}
SWE-Agent integrates LLM-based reasoning with specialized Agent-Computer Interfaces for automated software engineering~\cite{yang2024swe}. It provides custom commands for code editing, searching, and navigation optimized for LLM comprehension, achieving 12.5\% resolution rate on SWE-bench \cite{jimenez2023swe} benchmarks (3.3$\times$ improvement over baselines). The computation pattern of SWE-Agent primarily involves iterative code refinement and specialized interfaces, which appear in Devin \cite{devin-ai} and Claude code \cite{claude-code} as well.

\subsection{RAG (Haystack)}

We implement ENNS retrieval from C4 \cite{dodge2021documenting} document corpus (115 GB english variant). In a controlled QA-RAG study \cite{quinn2025accelerating}, ENNS outperform Approximate Nearest Neighbor Search (ANNS) in generation accuracy by 22.6–53.4\% at K=1 (document count) and 13.6-45.2\% at K=16 across FiDT5, Llama-3-8B, and Llama-3-70B models. Moreover, the study also concluded that ENNS dominates the throughput-accuracy pareto-frontier as compared to ANNS. Following the same setting used in the study, we choose CPU-based FAISS \cite{douze2024faiss} retrieval due to large document size, far exceeding the GPU memory.

We implement this RAG workload using Haystack \cite{deepset-haystack} which provides a production-ready framework for building RAG pipelines and question-answering systems. It implements directed multigraph architectures with modular components for retrieval (BM25, dense embeddings) and generation, achieving F1=82.91 on SQuAD 2.0 \cite{rajpurkar2018know} benchmarks. Haystack computation pattern primarily involves pipeline orchestration and hybrid retrieval, which appear in LlamaIndex \cite{llamaindex} and  Semantic Kernel \cite{microsoft-no-date} as well.

\subsection{ChemCrow}
ChemCrow \cite{bran2023chemcrow} augments LLMs with specialized chemistry tools for scientific research automation. ChemCrow integrates 18 expert-designed tools spanning reaction prediction, molecular analysis, and safety assessment, using ReAct-style reasoning chains. It outperforms GPT-4 by 4.4/10 points in expert evaluations and achieves 100\% success rate on synthesis tasks. The computation pattern of ChemCrow primarily involves domain-specific tool integration and ReAct reasoning, which also appear in GeoGPT \cite{zhang2023geogpt}.

\subsection{Web-Augmented Agent (LangChain)}

We choose a web-augmented agentic pipeline (web search $\to$ summarization $\to$ LLM inference) inspired by the web search feature of popular chatbots \cite{gemini2026, chatgpt2026}. The summarizer reduces the prompt length and parses factual information from web documents. We chose a CPU-based LexRank summarizer \cite{erkan2004lexrank} compared to an LLM-based summarizer because of two reasons. \textit{\underline{First}, Hallucinations: }A study \cite{maynez2020faithfulness} shows that on the XSum benchmark \cite{narayan2018don}, 73–79\% of model summaries contained at least one hallucination and the best system still had 64\% extrinsic hallucinations. \textit{\underline{Second}, Domain Accuracy: }The accuracy of LexRank-based summarizer is within 0.05 ROUGE-1 of LLM-based summarizer for DUC-2004 benchmark and even surpasses for legal benchmark, such as BillSum \cite{giarelis2023abstractive}. 

We implement this agent using a popular framework called LangChain \cite{mavroudis2024langchain} which facilitates composable agent development through modular chains and graph-based orchestration. LangChain consists of core abstractions for tool calling, memory management, and stateful multi-agent coordination. The computation pattern of LangChain primarily involves chain composition and stateful orchestration, that appear in CrewAI \cite{crewai_moura_2025} and AutoGen \cite{wu2024autogen} as well.

\section{Multi-processing and Multi-threading}
\label{appendix:section-multiprocessing}
\begin{figure}
\centering
    \includegraphics[width=0.9\columnwidth]
    {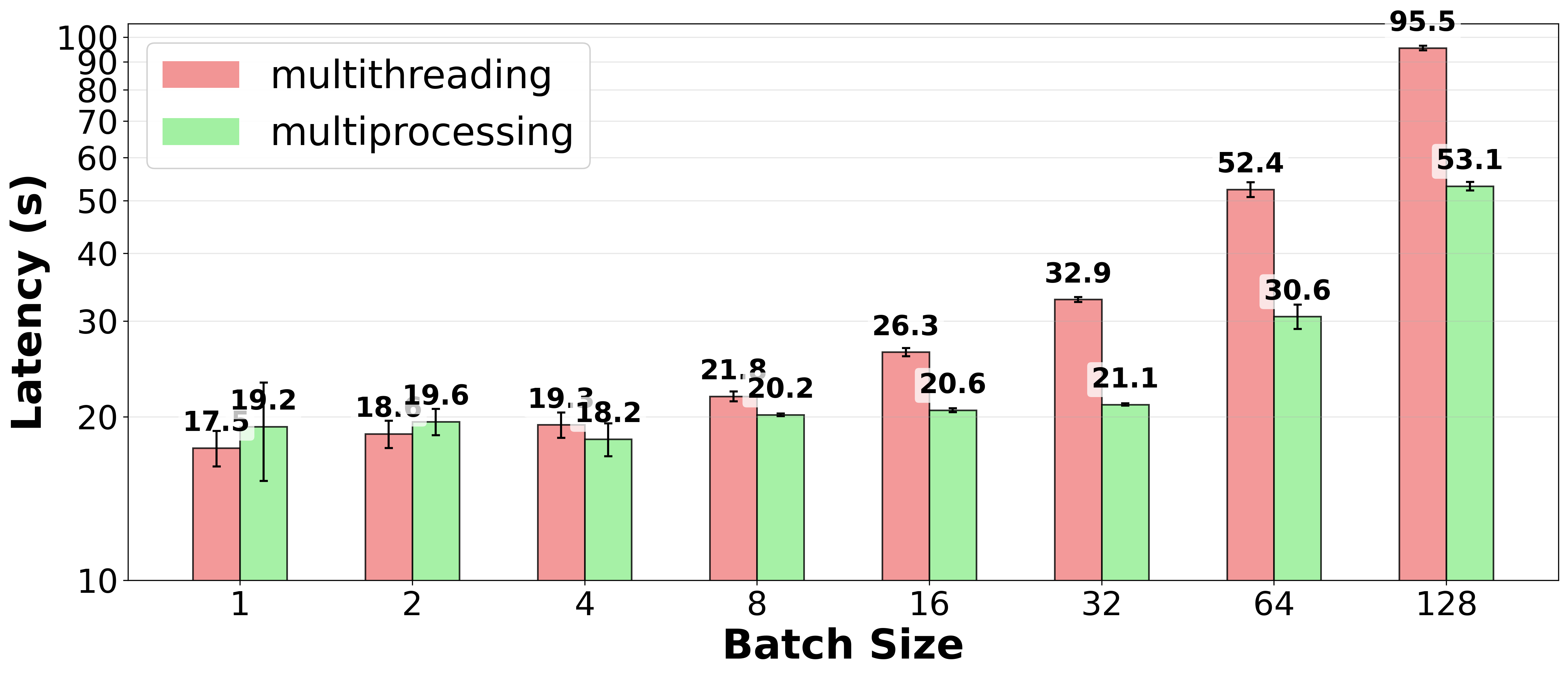}
    \caption{Comparison of multi-processing and multi-threading with single core baseline for Web-Augmented workload on \texttt{Sys 2}.}
    \vspace{-3mm}
    \label{fig:multiprocessing}
\end{figure}

We compare the latency of multi-processing and multi-threading for the LangChain workload across varying batch sizes in \autoref{fig:multiprocessing}. LangChain’s built-in \emph{Runnable.batch} API processes multiple inputs concurrently within one process implemented via a thread pool (multi-threading). Multi-processing launches N independent Python processes (each serving batch size 1) using the shell background operator `\&', thereby achieving coarse-grained parallelism across CPU cores and sidestepping single-process Global Interpreter Lock (GIL) limitations. Moreover, multi-processing also mitigates the synchronization overheads incurred by multi-threading. As observed, performance between the two approaches is relatively comparable at low batch sizes ($\le 4$). However, as the batch size scales, the synchronization overheads and GIL bottleneck associated with multi-threading severely degrade performance. Consequently, at a batch size of 128, multi-processing achieves an approximate 1.8× speedup over multi-threading (53.1s compared to 95.5s).

\section{MAS Algorithm}
\label{appendix:mas_algorithm}

\begin{algorithm}[t]
\caption{Mixed Agentic Scheduler (MAS): type-aware elastic admission with per-class waiting queues}
\label{alg:mas}
\begin{algorithmic}[1]
\State \textbf{Active concurrency queues:} execution queues $Q_{\mathrm{CPU}}$, $Q_{\mathrm{GPU}}$; shared reserve queue $Q_{\mathrm{res}}$
\State \textbf{Waiting queues:} per-class FCFS queues $W_{\mathrm{CPU}}$, $W_{\mathrm{GPU}}$
\State \textbf{Concurrency caps:} $E_{\mathrm{cap,CPU}}$, $E_{\mathrm{cap,GPU}}$, $E_{\mathrm{cap,shared}}$
\State \textbf{Round-robin pointer:} $next \in \{\mathrm{CPU}, \mathrm{GPU}\}$

\Procedure{TryAdmit}{$r$}
    \If{$r$ is CPU-heavy \textbf{and} $|Q_{\mathrm{CPU}}| < E_{\mathrm{cap,CPU}}$}
        \State enqueue $r$ into $Q_{\mathrm{CPU}}$
        \State \Return \textbf{true}
    \ElsIf{$r$ is GPU-heavy \textbf{and} $|Q_{\mathrm{GPU}}| < E_{\mathrm{cap,GPU}}$}
        \State enqueue $r$ into $Q_{\mathrm{GPU}}$
        \State \Return \textbf{true}
    \ElsIf{$|Q_{\mathrm{res}}| < E_{\mathrm{cap,shared}}$}
        \State enqueue $r$ into $Q_{\mathrm{res}}$ \Comment{shared overflow queue}
        \State \Return \textbf{true}
    \Else
        \State \Return \textbf{false}
    \EndIf
\EndProcedure

\Procedure{OnArrival}{$r$}
    \If{\textproc{TryAdmit}($r$) = \textbf{false}}
        \If{$r$ is CPU-heavy}
            \State enqueue $r$ into $W_{\mathrm{CPU}}$
        \Else
            \State enqueue $r$ into $W_{\mathrm{GPU}}$
        \EndIf
    \EndIf
\EndProcedure

\Procedure{DrainWaiting}{}
    \If{$next = \mathrm{CPU}$}
        \State $first \gets \mathrm{CPU}$; $second \gets \mathrm{GPU}$
    \Else
        \State $first \gets \mathrm{GPU}$; $second \gets \mathrm{CPU}$
    \EndIf

    \If{$W_{first}$ is not empty \textbf{and} \textproc{TryAdmit}(head of $W_{first}$)}
        \State dequeue head of $W_{first}$
        \State $next \gets second$
    \EndIf

    \If{$W_{second}$ is not empty \textbf{and} \textproc{TryAdmit}(head of $W_{second}$)}
        \State dequeue head of $W_{second}$
        \State $next \gets first$
    \EndIf
\EndProcedure

\Procedure{OnCompletion}{}
    \State \textproc{DrainWaiting}()
\EndProcedure

\Statex
\Statex \textbf{Dispatch policy:}
\Statex CPU-heavy requests are served from $Q_{\mathrm{CPU}}$.
\Statex GPU-heavy requests are served from $Q_{\mathrm{GPU}}$.
\Statex Requests in $Q_{\mathrm{res}}$ are served in FCFS order.
\Statex Waiting queues are FCFS within class and are drained in round-robin order on each completion.
\Statex Elastic admission is always attempted before reserve admission.
\end{algorithmic}
\end{algorithm}
\normalsize

\hyperref[alg:mas]{Algorithm 1} summarizes the admission logic of MAS. MAS maintains two execution queues, $Q_{\mathrm{CPU}}$ and $Q_{\mathrm{GPU}}$, which hold CPU-heavy and GPU-heavy requests that have already been admitted and are allowed to make progress in the system. The number of requests that can be placed in these queues is limited by the request-type-specific concurrency caps $E_{\mathrm{cap,CPU}}$ and $E_{\mathrm{cap,GPU}}$. It also maintains a shared reserve execution queue $Q_{\mathrm{res}}$ with capacity $E_{\mathrm{cap,shared}}$, which absorbs overflow from either request-type when its corresponding execution queue is full. On arrival, a request $r$ is first classified as CPU-heavy or GPU-heavy and is admitted to its matching execution queue whenever capacity is available; otherwise MAS attempts reserve admission. If both options are full, the request is placed into a per-request waiting queue, $W_{\mathrm{CPU}}$ or $W_{\mathrm{GPU}}$, each of which preserves FCFS order within that request-type. Upon every completion, MAS revisits the waiting queues and attempts admission from their heads, always prioritizing the request-type-specific execution queues before the shared reserve queue. To avoid cross-request head-of-line blocking while preserving a simple and race-free policy, MAS drains $W_{\mathrm{CPU}}$ and $W_{\mathrm{GPU}}$ in round-robin order rather than in parallel: this allows a blocked CPU-heavy request to no longer stall a GPU-heavy request behind it, while also avoiding unsynchronized concurrent access to shared admission state, particularly the shared reserve queue $Q_{\mathrm{res}}$ and its cap $E_{\mathrm{cap,shared}}$. Overall, this design preserves request-type-aware elasticity under normal load, uses the shared reserve only for overflow, and reduces cross-request interference while maintaining fair and work-conserving admission.

\end{document}